\documentclass[preprint,12pt]{elsarticle}

\usepackage{amssymb}
\usepackage{amsthm}
\usepackage{lineno}
\usepackage{bm}
\usepackage[colorlinks=true, pdfauthor=author]{hyperref}
\usepackage{wrapfig}
\usepackage{booktabs} 
\usepackage{algorithm}
\usepackage{algorithmic}
\usepackage{graphicx}
\usepackage{amsmath}
\usepackage{amssymb}
\usepackage{amsfonts}
\usepackage{bbding}
\usepackage[table]{xcolor}
\usepackage{multirow}
\usepackage{graphicx}
\usepackage{mathrsfs}
\usepackage{colortbl}
\usepackage{subfigure}
\usepackage{indentfirst}
\usepackage{ulem}
\usepackage{url}
\usepackage{pifont}
\usepackage{verbatim}
\usepackage{rotating}
\usepackage{adjustbox}
\usepackage{wrapfig}
\usepackage{setspace}
\usepackage{colortbl}
\usepackage[table]{xcolor}

\graphicspath{{./images/}}

\journal{arXiv}
\begin{document}

\begin{frontmatter}



\title{TIME: TabPFN-Integrated Multimodal Engine for Robust Tabular-Image Learning}

\author[first]{Jiaqi Luo}
\ead{jqluo@suda.edu.cn}
\author[second]{Yuan Yuan}
\ead{y.yuan@dukekunshan.edu.cn}
\author[second]{Shixin Xu\corref{cor1}}
\cortext[cor1]{Corresponding author}
\ead{shixin.xu@dukekunshan.edu.cn}
\affiliation[first]{organization={School of Mathematical Sciences, Soochow University},
            addressline={No.1 Shizi Street}, 
            city={Suzhou},
            postcode={215006}, 
            state={Jiangsu Province},
            country={China}}
\affiliation[second]{organization={ Zu Chongzhi Center, Duke Kunshan University},
            addressline={No.8 Duke Avenue}, 
            city={Kunshan},
            postcode={215000}, 
            state={Jiangsu Province},
            country={China}}

\begin{abstract}

Tabular-image multimodal learning, which integrates structured tabular data with imaging data, holds great promise for a variety of tasks, especially in medical applications. Yet, two key challenges remain: (1) the lack of a standardized, pretrained representation for tabular data, as is commonly available in vision and language domains; and (2) the difficulty of handling missing values in the tabular modality, which are common in real-world medical datasets.
To address these issues, we propose the \textbf{T}abPFN-\textbf{I}ntegrated \textbf{M}ultimodal \textbf{E}ngine (\textbf{TIME}), a novel multimodal framework that builds on the recently introduced tabular foundation model, TabPFN. TIME leverages TabPFN as a frozen tabular encoder to generate robust, strong embeddings that are naturally resilient to missing data, and combines them with image features from pretrained vision backbones.
We explore a range of fusion strategies and tabular encoders, and evaluate our approach on both natural and medical datasets. Extensive experiments demonstrate that TIME consistently outperforms competitive baselines across both complete and incomplete tabular inputs, underscoring its practical value in real-world multimodal learning scenarios.

\end{abstract}




\end{frontmatter}


\section{Introduction}
\label{s:intro}
Multimodal learning has emerged as a powerful paradigm for integrating diverse data sources to enhance learning and decision-making across a wide range of domains \cite{baltruvsaitis2018multimodal}. Among the many forms of multimodal integration, tabular-image multimodal learning plays a uniquely important role, especially in clinical and biomedical applications \cite{acosta2022multimodal, huang2020fusion}. In such settings, structured tabular data such as laboratory test results often coexist with unstructured imaging data like X-rays. Effectively combining these two complementary modalities can yield richer and more context-aware representations, leading to more accurate diagnoses and clinical decisions \cite{duanmu2020prediction, borsos2024predicting, vale2021long}. As such, tabular-image multimodal learning lies at the forefront of modern AI applications in healthcare and beyond.

A central enabler of recent success in multimodal learning is the use of pretrained foundation encoders. Vision models such as ResNet \cite{he2016deep} and ViT \cite{dosovitskiy2020image}, and language models like BERT \cite{devlin2019bert}, have demonstrated that pretrained representations not only accelerate convergence and reduce the need for large labeled datasets but also generalize well across tasks. 
However, structured tabular data remains an outlier in this landscape. Tabular data are inherently heterogeneous—combining numerical, categorical, and often missing values—making it difficult to standardize their representation. 

Critically, there is no widely adopted pretrained tabular encoder, leaving most existing multimodal pipelines to train tabular encoders from scratch for each task \cite{jiang2024tabular, holste2021end}. This undermines generalization, particularly in small-data settings, and increases computational burden.
Moreover, existing approaches commonly assume that tabular inputs are complete, relying on basic imputation techniques such as mean or median filling. These methods risk distorting the feature distribution or masking informative patterns of missingness, which are especially prevalent in real-world clinical datasets \cite{du2024tip}. The inability to natively handle missing data further limits the robustness and applicability of these models in practical deployments.

To bridge this gap, recent work has turned to contrastive learning to align tabular and image representations \cite{du2024tip, hager2023best, huang2023multimodal}. While promising, these methods are typically trained on a per-task basis, requiring extensive paired data for each application. Once trained, the resulting encoders are often specific to the original task, limiting their reusability across domains. Moreover, contrastive methods generally do not account for missing values during training, further restricting their applicability in domains like healthcare, where data is often incomplete.

To address these limitations, we propose a new multimodal learning framework that integrates TabPFN, a pretrained tabular foundation model capable of zero-shot generalization and handling missing values natively. Trained on millions of synthetic tasks, TabPFN uses a transformer-based architecture and in-context learning to produce accurate tabular predictions without gradient updates. In our framework, we treat TabPFN as a frozen tabular encoder and combine its embeddings with features extracted from a pretrained vision backbone. This design introduces both robustness to missingness and the benefits of tabular pretraining into multimodal learning. 
We also explore various fusion strategies and model combinations, and demonstrate through extensive experiments on natural and medical datasets that our method outperforms existing strong baselines across settings with complete and incomplete tabular inputs. Our findings underscore the practical value of leveraging pretrained tabular representations to build more generalizable and resilient multimodal systems.

The main contributions of this paper are as follows:
\begin{itemize}
    \item We introduce a novel multimodal learning framework that explicitly addresses two core challenges: the lack of universal pretrained tabular representations and the inability to handle missing values in structured data
    \item We are the first to incorporate TabPFN, a transformer-based tabular foundation model, into a multimodal pipeline, enabling robust and zero-shot tabular embeddings under complete and incomplete conditions.
    \item We conduct extensive experiments on five real-world datasets, demonstrating that our approach significantly outperforms existing baselines across various fusion strategies and settings.
\end{itemize}


\section{Related work}
\label{s:rel}
\subsection{Tabular Machine Learning}
Tabular data, commonly encountered in areas such as finance, healthcare, and scientific research, highlights the importance of precise predictions. In practical applications, decision tree models \cite{chen2016xgboost,ke2017lightgbm,prokhorenkova2018catboost} are favored for tabular data analysis due to their interpretability, ability to handle missing values, and cheap computational cost. However, the lack of flexibility  also limits tree-based methods to valuable tasks like transfer learning \cite{wang2022transtab} and mutlimodal learning \cite{du2024tip, hager2023best}.

Deep learning models have become popular in the analysis of tabular data \cite{borisov2022deep}, as their flexibility addresses the limitations of decision trees \cite{luo2024ncart}, enabling valuable applications like self-supervised learning \cite{yoon2020vime, bahri2021scarf} and generative modeling \cite{xu2019modeling}. 
However, most existing studies do not address missing values, which is a common problem in practice, especially in healthcare. Most methods will fill missing positions and only a few models have introduced the mechanisms to overcome these limitations.
TabTransformer \cite{huang2020tabtransformer} can only handle the missing values existing in the test data, it uses the average learned embeddings over all classes or the embedding for the class of missing value to fill in the missing values.
Self-Attention and Intersample Attention Transformer (SAINT) \cite{somepalli2021saint} can also handle missing scenarios, it borrows the corresponding features from other similar data samples. 
Recently, TabPFN \cite{hollmann2025accurate}, a foundation model based on in-context learning, has been presented for small-sized tabular tasks. The model is a transformed-based pretrained model that can handle missing values without preprocessing, which inspires us to adapt it for image-tabular learning with missingness.

\subsection{Tabular-Image Multimodal Learning}
In the context of tabular-image multimodal learning, the goal is to integrate structured tabular data and unstructured image data to build more robust models. Existing approaches typically fall under two broad categories: supervised learning and self-supervised learning.

\paragraph{Supervised Learning}

In supervised multimodal learning, a common approach is to use separate encoders to extract modality-specific features: a vision encoder for images and a tabular encoder for structured data. The resulting embeddings are then fused to generate the final prediction. The image encoder is usually a pretrained model such as ResNet \cite{he2016deep} or ViT \cite{dosovitskiy2020image}, while the tabular encoder typically employs specialized deep learning architectures such as MLP or Transformer-based models \cite{borisov2022deep}, often trained from scratch. Several studies have demonstrated that integrating complementary signals from these modalities significantly enhances performance on complex tasks.

For example, CHARMS \cite{jiang2024tabular} uses a Transformer to extract tabular representations and transfers this knowledge to improve visual predictions, but it only utilizes the image encoder for the prediction, neglecting the wealth of information in the tabular data for decision-making. CLIP-Lung \cite{lei2023clip} incorporates a pretrained language model to inject clinical textual knowledge into the image encoder, thereby enhancing classification performance. Nodule-CLIP \cite{sun2024nodule} similarly improves visual reasoning by first segmenting 3D lung nodules using a U-Net, reducing irrelevant information and focusing the model on the relevant anatomical regions.
Spasov et al. \cite{spasov2019parameter} propose a multi-task framework that integrates MRI images and clinical data to jointly predict both the progression from mild cognitive impairment (MCI) to Alzheimer’s disease (AD) and the classification of AD versus healthy controls. Liu et al. \cite{liu2023functional} combine Support Vector Regression with a 3D ResNet to improve predictions in acute ischemic stroke. Multi-TransSP \cite{zheng2022multi} introduces a Transformer-based framework that fuses CNN-extracted image features with Transformer-processed tabular data for survival prediction using CT images and structured clinical variables. Xue et al. \cite{xue2024ai} unify diverse modalities into fixed-length embeddings and use a Transformer to identify etiologies contributing to dementia.
Additionally, several works focus explicitly on improving fusion strategies. DAFT (Dynamic Affine Feature Map Transform) \cite{wolf2022daft} is a general-purpose module that dynamically conditions CNN feature maps on both image and tabular inputs, enhancing or suppressing high-level visual concepts accordingly. Tang et al. \cite{tang2022improving} propose models that jointly leverage structured and unstructured data for lung nodule classification. Holste et al. \cite{holste2021end} evaluate various fusion strategies across different stages in an end-to-end architecture that integrates imaging and non-imaging data.

Although these studies demonstrate that integrating complementary signals can significantly enhance performance across diverse tasks and dataset sizes, they overlook the issue of missing values—an important limitation in real-world applications. This motivates us to explore multimodal learning under conditions of incomplete data. To the best of our knowledge, we are the first to address tabular-image multimodal learning with missing values in a supervised setting.

\paragraph{Self-supervised Learning}
Self-supervised learning techniques for multimodal learning often rely on contrastive learning, where the objective is to bring matching tabular–image pairs closer in the embedding space while pushing apart non-matching pairs. The resulting pretrained encoders can then be fine-tuned or directly applied to downstream supervised tasks.

MMCL \cite{hager2023best} is the first work to apply contrastive learning in the tabular-image multimodal setting. It learns representations from large-scale image-tabular pairs and demonstrates significant gains on both natural and medical datasets compared to purely supervised or self-supervised vision methods.
Huang \cite{huang2023multimodal} introduces a novel tabular attention module and also uses contrastive learning to enhance Alzheimer’s disease prediction.
TIP \cite{du2024tip} is the first to address missingness in downstream tabular data and disparities between tabular and image modalities. It incorporates a masked tabular reconstruction task to make tabular representations robust to missing values and proposes an tabular-image matching strategy, ultimately achieving strong performance across multiple supervised, self-supervised, and multimodal benchmarks.

While these methods generate excellent encoders, they only focus on classification and are trained case by case, \textit{i.e.}, where contrastive learning is conducted anew for each task, and the pretrained model is then applied only to the same downstream task, which are computationally expensive, limiting their usefulness in small-sized datasets and more general-purpose settings.



\section{Methodology}
This section describes our proposed multimodal learning framework for integrating structured tabular data and unstructured image data for classification and regression tasks. We first revisit the pretrained TabPFN model used to extract tabular embeddings. We then present the overall architecture of our multimodal learning framework, which combines features from both modalities via various fusion strategies.

\subsection{Revisiting Pretrained TabPFN}

\paragraph{TabPFN V1}  
Tabular Prior-data Fitted Network (TabPFN) \cite{hollmann2022tabpfn} is a transformer-based model trained offline to approximate Bayesian inference on synthetic tabular datasets. It integrates priors from Bayesian Neural Networks and Structural Causal Models, enabling the model to capture complex feature dependencies and potential causal structures.
A key feature of TabPFN is its ability to perform inference on new datasets using a single forward pass, avoiding the need for model re-training. This is achieved via in-context learning, where the model conditions on a set of labeled examples and produces predictions for new inputs without parameter updates. 
We now outline the prediction process of TabPFN; for further details, we refer the reader to the original paper and the
official implementation \footnote{https://github.com/PriorLabs/TabPFN/tree/main}.

Let the training set be $\mathcal{D}_{\text{train}} = \{ (\mathbf{X}_i, y_i) \}_{i=1}^{N}$, where $\mathbf{X}_i \in \mathbb{R}^{d}$ denotes a tabular feature vector and $y_i \in \{1, \ldots, C\}$ is the class label.
TabPFN requires the full training set \( \mathcal{D}_{\text{train}} \) to be loaded at inference time to condition its predictions. Given a test instance \( \mathbf{X}_{\text{test}} \), TabPFN outputs the posterior predictive distribution (PPD):
\begin{equation}
    p(y_{\text{test}} \mid \mathbf{X}_{\text{test}}, \mathcal{D}_{\text{train}}).
\end{equation}

\paragraph{TabPFN V2}  
Hollmann et al. \cite{hollmann2025accurate} extend TabPFN V1 with the following enhancements:
\begin{itemize}
    \item The classification capability has been improved and expanded to support regression tasks. 
    \item Natively supports missing values and outliers, making the processing of datasets more efficient and accurate. 
    \item With the help of flash attention, the memory and computing requirements during training and inference are optimized. 
    \item Designed for small- to medium-sized datasets ($\le$10000 samples, $\le$500 features, $\le$10 classes).
\end{itemize}

In our framework, we use TabPFN v2 to extract tabular embeddings by accessing the internal encoder output. Let $f(\cdot)$ denote this embedding function. Then, for any input $\mathbf{X_*}$, the tabular embedding is given by:
\begin{equation}
    \mathbf{E} = f(\mathbf{X_*}, \mathcal{D}_{\text{train}}),
\end{equation}
where $\mathbf{X_*}$ may belong to training, validation, or test sets. In TabPFN v2 paper, the embedding dimension is 192.

\subsection{TabPFN-Integrated Multimodal Engine}
\begin{figure}[!ht]
    \centering
    \includegraphics[width=0.95\linewidth]{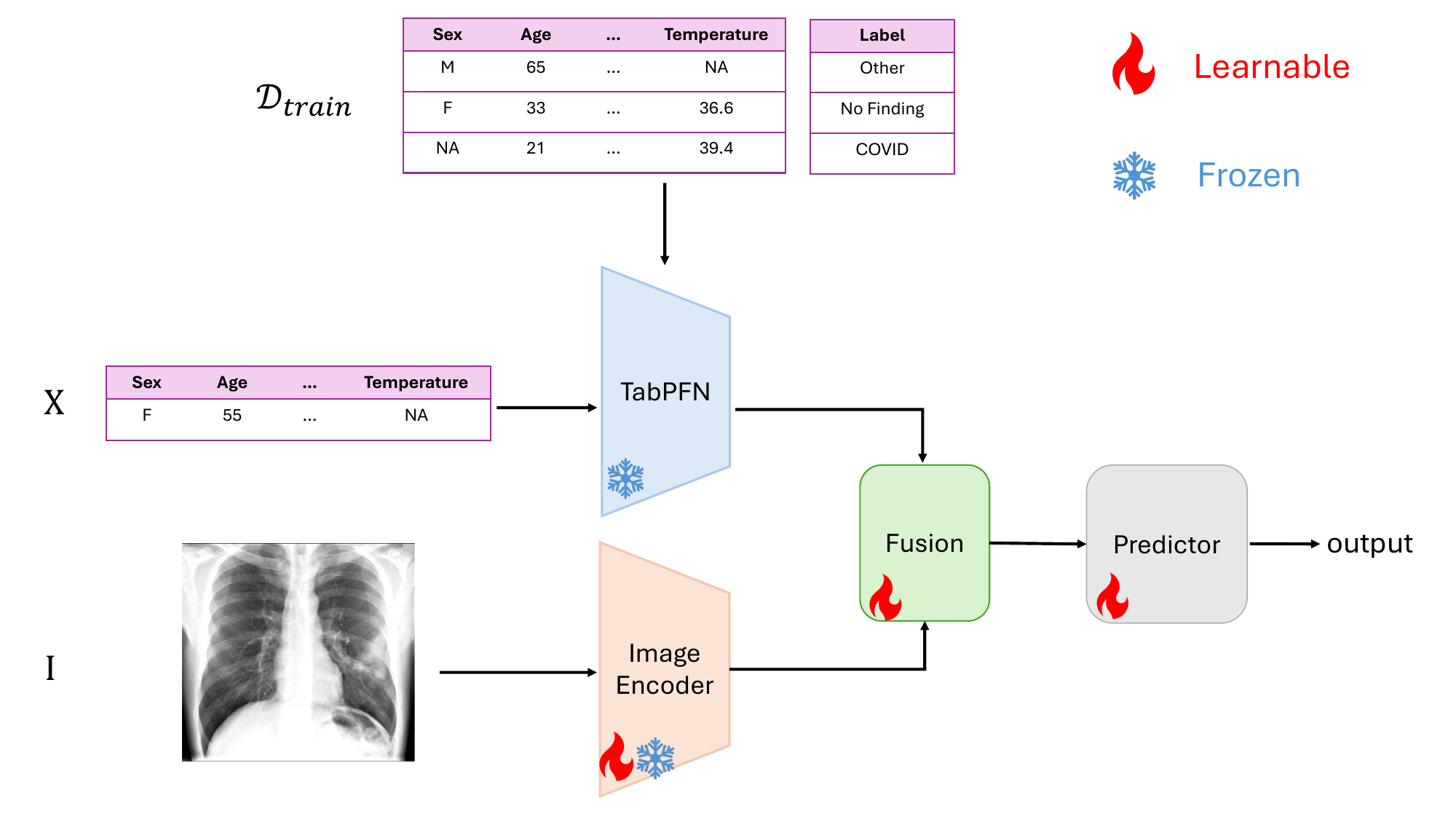}
    \caption{TIME model architecture. Blocks marked with red flames are trainable, while blocks marked with blue snowflakes are frozen during training. The image encoder can be either trainable or frozen.}
    \label{f.framework}
\end{figure}

Our proposed model, which is shown in Fig.~\ref{f.framework}, consists of four components: (1) tabular embedding extraction via TabPFN, (2) image embedding extraction, (3) multimodal feature fusion, and (4) prediction: classification or regression.


In our design, the TabPFN encoder is kept frozen, as it is trained offline using in-context learning. At inference time, TabPFN conditions on the entire training set and outputs predictions for test instances in a single forward pass, without requiring parameter updates.
As such, TabPFN is not intended to be fine-tuned on downstream tasks. Freezing it preserves its generalization ability and aligns with its original training and test paradigm. In contrast, the ResNet encoder, which is trained in a supervised manner, can either be frozen or fine-tuned depending on the availability of data and the desired adaptability to the downstream task.

\paragraph{Tabular Embedding} Given tabular input $\mathbf{X} \in \mathbb{R}^{d}$ and training set $\mathcal{D}_{\text{train}}$, the tabular embedding is extracted via TabPFN as:
\begin{equation}
\label{e.tabemb}
    \mathbf{E}_T = f_{\text{TabPFN}}(\mathbf{X}, \mathcal{D}_{\text{train}}).
\end{equation}

\paragraph{Image Embedding} Given an image $\mathbf{I} \in \mathbb{R}^{H \times W \times C}$, we use a ResNet \cite{he2016deep} encoder $f_{\text{ResNet}}$ to extract image features. The encoder may be either frozen or fully-tuned. The image embedding is:
\begin{equation}
\label{e.imgemb}
    \mathbf{E}_I = f_{\text{ResNet}}(\mathbf{I}).
\end{equation}

\paragraph{Multimodal Fusion} 
We explore several fusion strategies to combine the tabular and image embeddings:
\begin{itemize}
    \item Concatenation: $\mathbf{Z} = [\mathbf{E}_{T}; \mathbf{E}_I].$
    \item Element-wise Sum: $\mathbf{Z} = \mathbf{E}_{T} + \mathbf{E}_I.$
    \item Element-wise Maximum: $\mathbf{Z} = \max(\mathbf{E}_{T}, \mathbf{E}_I)$, where the max operation is applied element-wise.
    \item Dynamic Affine Feature Map Transform \cite{wolf2022daft}: $\mathbf{Z} = DAFT(\mathbf{E}_{T}, \mathbf{E}_I)$.
\end{itemize}

For element-wise operations, both embeddings are projected into a common dimension $k$ using linear layers:
\begin{equation*}
    \mathbf{E}_T' = \mathbf{W}_T \mathbf{E}_T, ~~\mathbf{E}_I' = \mathbf{W}_I \mathbf{E}_I,
\end{equation*}
where $\mathbf{E}_T', \mathbf{E}_I' \in \mathbb{R}^{k}$, and $\mathbf{W}_T$ and $\mathbf{W}_I$ are learnable.

\paragraph{Prediction} 
The fused representation is passed through a simple linear classifier:
\begin{equation}
    \hat{y} = softmax(\mathbf{W} \mathbf{Z} + \mathbf{b}),
\end{equation}
or a simple linear regressor:
\begin{equation}
    \hat{y} = \mathbf{W} \mathbf{Z} + \mathbf{b},
\end{equation}
where $\mathbf{W}$ and $\mathbf{b}$ are learnable weights.

\section{Experiments}
\label{s:exps}
\subsection{Experiments Setup}
\paragraph{Datasets}
We conduct experiments on five publicly available datasets from Kaggle, including four classification tasks \footnote{https://www.kaggle.com/competitions/petfinder-adoption-prediction/data} \footnote{https://www.kaggle.com/datasets/awsaf49/cbis-ddsm-breast-cancer-image-dataset} \footnote{https://www.kaggle.com/datasets/aaryapatel/covid19-chest-xray} \footnote{https://www.kaggle.com/datasets/mahdavi1202/skin-cancer} and one regression task \footnote{https://www.kaggle.com/datasets/denozavrus/paintings-price-prediction/data}. Among them, one is a natural image dataset, one is a painting dataset, and the remaining three are medical image datasets. Notably, all three medical datasets contain missing values in the tabular modality.
For datasets with predefined training/test splits, we use the original partitions. For the others, we randomly split each dataset, allocating 80\% of the instances for training and the remaining 20\% for testing. From the training set, we further reserve 20\% as a validation set for model selection and early stopping. Detailed descriptions of the datasets can be found in \ref{a.data}.

\paragraph{Implementation Details}
We use ResNet-50 \cite{he2016deep} as the image encoder, resizing all input images to 256$\times$256 pixels. We train the ResNet using linear probing, which only tunes the linear classifier/regressor, and fully fine-tuning, which trains all parameters.
For comparative tabular encoders, we include a standard MLP \cite{hager2023best} and NCART \cite{luo2024ncart}. Since these models do not support missing values, we apply median imputation before training.
We use accuracy (Acc.) for classification tasks and mean squared error (MSE) for regression to evaluate performance. All models are implemented in PyTorch and trained on a single GPU with 24 GB memory. We train each model for 100 epochs using the AdamW optimizer, with an initial learning rate of 1e-3, batch size of 64, and a learning rate decay factor of 0.9 every 20 epochs. To ensure statistical robustness, all experiments are repeated with five different random seeds, and we report the mean and standard deviation of the results \footnote{The code will be released upon acceptance.}.

\subsection{Performance Comparison on Complete Data}

\begin{sidewaystable}
\renewcommand\arraystretch{1.3}
\centering
\caption{Performance scores (Mean$\pm$std.) of different methods on five complete datasets. The \underline{\textbf{bold}} indicates the top result. \textcolor{red}{\ding{88}} indicates full fine-tuning, \textit{i.e.}, all parameters are trainable. \textcolor{blue}{\ding{100}} denotes linear probing, \textit{i.e.}, the image encoder is frozen and only the linear classifier/regressor is updated during training. The gray-colored “Mean” lines indicate the average scores of each model across different fusion strategies.}
\label{T.results}
\begin{adjustbox}{width=\textwidth}
\begin{tabular}{l|cc|cc|cc|cc|cc}
\toprule[2pt]
\multirow{2}{*}{Models} & \multicolumn{2}{c|}{Adoption (Acc. $\uparrow$)} & \multicolumn{2}{c|}{Breast Cancer (Acc. $\uparrow$)} & \multicolumn{2}{c|}{Covid-19 (Acc. $\uparrow$)} & \multicolumn{2}{c|}{Skin Cancer (Acc. $\uparrow$)} & \multicolumn{2}{c}{Painting (MSE $\downarrow$)}\\
      &  \Large\textcolor{red}{\ding{88}}  &   \Large\textcolor{blue}{\ding{100}}  &   \Large\textcolor{red}{\ding{88}}   &   \Large\textcolor{blue}{\ding{100}}   &   \Large\textcolor{red}{\ding{88}}   &   \Large\textcolor{blue}{\ding{100}}  &   \Large\textcolor{red}{\ding{88}}  &   \Large\textcolor{blue}{\ding{100}}  & \Large\textcolor{red}{\ding{88}}  &   \Large\textcolor{blue}{\ding{100}}  \\
\midrule[1.5pt]

ResNet-50  &  29.91\small$\pm$0.57   &   34.09\small$\pm$0.58   &  60.92\small$\pm$2.09 &  59.26\small$\pm$1.15  &  81.46\small$\pm$2.06  &  76.95\small$\pm$0.60  &   67.96\small$\pm$1.66   &   59.61\small$\pm$1.15  &  20.02\small$\pm$0.26  &  19.63\small$\pm$0.04   \\

\midrule[1pt]
MLP-Cat  &  29.32\small$\pm$0.38    &   34.50\small$\pm$0.36   &  66.87\small$\pm$2.42  &  70.74\small$\pm$1.16  &  81.95\small$\pm$1.42  &  84.02\small$\pm$1.41    &  74.74\small$\pm$1.67   & 74.70\small$\pm$0.53   &  18.47\small$\pm$0.30  &  18.64\small$\pm$0.29       \\

MLP-Sum  &  29.51\small$\pm$0.56    &   33.57\small$\pm$0.79   &  67.67\small$\pm$3.17  &  69.69\small$\pm$0.74  &  81.59\small$\pm$2.26  & 78.54\small$\pm$1.70    &  75.09\small$\pm$1.46   &  72.48\small$\pm$1.35    &  18.67\small$\pm$0.42  &  18.83\small$\pm$0.17     \\

MLP-Max  &  29.33\small$\pm$1.10    &   34.63\small$\pm$0.53  &  65.21\small$\pm$2.66  &  70.61\small$\pm$1.39  &  80.85\small$\pm$1.37    & 79.27\small$\pm$0.86    &  75.43\small$\pm$1.09     &   72.57\small$\pm$0.35    &  17.67\small$\pm$0.26  &  17.36\small$\pm$0.16    \\

MLP-DAFT &  29.18\small$\pm$1.40    &   33.13\small$\pm$0.92   &  65.89\small$\pm$2.48   &  68.10\small$\pm$1.46   &  79.88\small$\pm$1.39    & 79.88\small$\pm$1.39  &  77.22\small$\pm$0.61     &  71.96\small$\pm$0.77    &  18.23\small$\pm$0.21  &  17.96\small$\pm$0.39     \\

\rowcolor{gray!20}
Mean &  29.34\small$\pm$0.86   & 33.96\small$\pm$0.65   & 66.41\small$\pm$2.68  & 69.78\small$\pm$1.19    &  81.07\small$\pm$1.61         &  80.33\small$\pm$1.51        &  75.62\small$\pm$1.21         &  72.92\small$\pm$0.75    &  18.26\small$\pm$0.29  &  18.20\small$\pm$0.25   \\

\midrule[1pt]
NCART-Cat  &  37.48\small$\pm$0.47    &   37.86\small$\pm$0.50   & 64.66\small$\pm$2.08   & 68.99\small$\pm$1.02   &  88.90\small$\pm$2.42     &  \underline{\textbf{89.76\small$\pm$1.24}}   & 78.83\small$\pm$0.98          &   75.96\small$\pm$1.46    &  17.94\small$\pm$0.43  &  17.80\small$\pm$0.22   \\

NCART-Sum  &  37.39\small$\pm$0.90    &  35.80\small$\pm$1.09  & 64.29\small$\pm$1.65   & 66.26\small$\pm$1.16   &  86.83\small$\pm$2.63    &  87.68\small$\pm$1.98    &  76.74\small$\pm$1.11   &   73.96\small$\pm$0.42    &  17.68\small$\pm$0.35  &  18.26\small$\pm$0.35    \\

NCART-Max  &  37.41\small$\pm$0.52    &  36.29\small$\pm$1.32  & 66.44\small$\pm$2.32   & 68.13\small$\pm$2.27   &  88.17\small$\pm$1.95       &  87.93\small$\pm$1.98        &  77.96\small$\pm$1.09         &  75.65\small$\pm$0.63     &  17.62\small$\pm$0.21  &  \underline{\textbf{16.69\small$\pm$0.28}}  \\

NCART-DAFT  &  37.29\small$\pm$0.86   & 35.95\small$\pm$0.43   & 67.61\small$\pm$2.43  & 69.69\small$\pm$1.15    &  88.66\small$\pm$2.36         &  88.66\small$\pm$2.36        &  79.35\small$\pm$0.82         &  74.52\small$\pm$1.75    &  17.49\small$\pm$0.22  &  17.79\small$\pm$0.38   \\

\rowcolor{gray!20}
Mean &  37.40\small$\pm$0.69   & 36.48\small$\pm$0.83   & 65.75\small$\pm$2.12  & 68.27\small$\pm$1.40    &  88.14\small$\pm$2.34         &  \underline{\textbf{88.29\small$\pm$1.76}}        &  78.22\small$\pm$1.00         &  75.02\small$\pm$1.07    &  17.68\small$\pm$0.30  &  \underline{\textbf{17.64\small$\pm$0.31}}   \\

\midrule[1pt]
TIME-Cat  &  39.15\small$\pm$1.00    &   \underline{\textbf{39.79\small$\pm$0.69}}  &  72.58\small$\pm$1.31  & \underline{\textbf{73.19\small$\pm$1.14}}    &  89.02\small$\pm$2.08      &  87.32\small$\pm$0.46     &  79.87\small$\pm$0.77     &   \underline{\textbf{79.22\small$\pm$0.58}}    &  17.25\small$\pm$0.13  &  17.27\small$\pm$0.04    \\

TIME-Sum  &  39.15\small$\pm$0.36    &   37.53\small$\pm$1.11  &  \underline{\textbf{73.68\small$\pm$1.94}}   & 72.64\small$\pm$0.98         &  88.66\small$\pm$1.47         &  87.07\small$\pm$1.24        &  \underline{\textbf{80.35\small$\pm$1.29}}     &   78.04\small$\pm$0.39    &  17.21\small$\pm$0.16  &  18.99\small$\pm$0.16    \\

TIME-Max  &  39.11\small$\pm$0.73   &   39.19\small$\pm$1.29 &  72.21\small$\pm$2.20   & 72.76\small$\pm$1.75   &  89.76\small$\pm$1.46         &  89.39\small$\pm$1.70        &  78.48\small$\pm$1.25         &   78.04\small$\pm$0.57     &  17.26\small$\pm$0.28  &  17.55\small$\pm$0.27   \\

TIME-DAFT  &  \underline{\textbf{39.60\small$\pm$0.53}}    &  37.99\small$\pm$0.76  &  72.33\small$\pm$3.51   & 72.21\small$\pm$1.09         &  \underline{\textbf{90.98\small$\pm$1.46}}         &  88.05\small$\pm$0.62    &  78.91\small$\pm$1.89         &  77.00\small$\pm$0.64    &  \underline{\textbf{17.21\small$\pm$0.14}}  &  17.01\small$\pm$0.72  \\

\rowcolor{gray!20}
Mean &  \underline{\textbf{39.25\small$\pm$0.68}}   & \underline{\textbf{38.63\small$\pm$0.96}}   & \underline{\textbf{72.70\small$\pm$2.24}}  & \underline{\textbf{72.70\small$\pm$1.24}}    &  \underline{\textbf{89.60\small$\pm$1.62}}         &  87.95\small$\pm$1.01        &  \underline{\textbf{79.40\small$\pm$1.30}}         &  \underline{\textbf{78.08\small$\pm$0.54}}    &  \underline{\textbf{17.23\small$\pm$0.18}}  &  17.70\small$\pm$0.30   \\

\bottomrule[2pt]
\end{tabular}
\end{adjustbox}

\end{sidewaystable}

\begin{figure}[!ht]
    \centering
    \includegraphics[width=\linewidth]{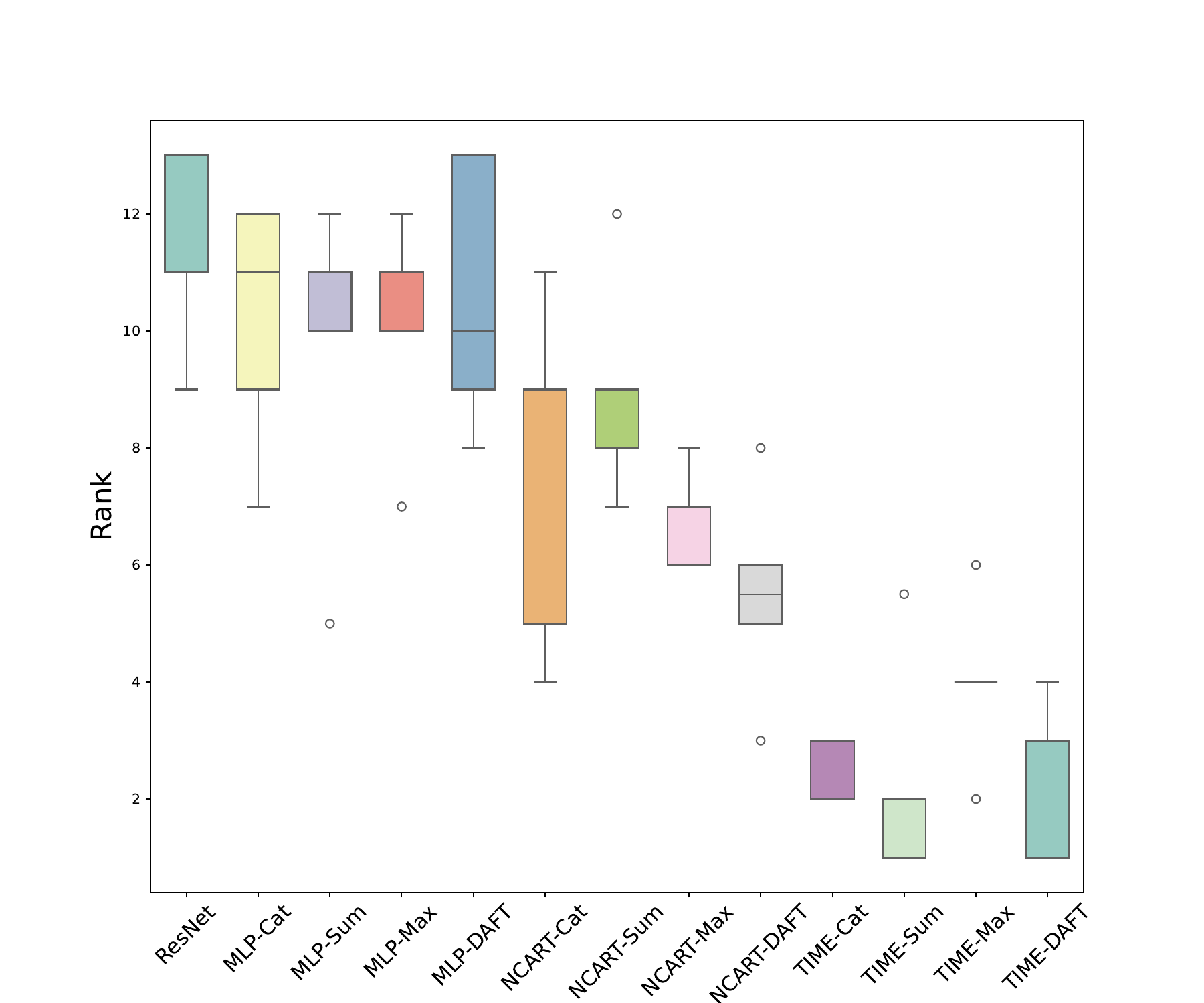}
    \caption{Rankings under the fully tuned setting on five datasets.}
    \label{f.fullyrank}
\end{figure}

\begin{figure}[!ht]
    \centering
    \includegraphics[width=\linewidth]{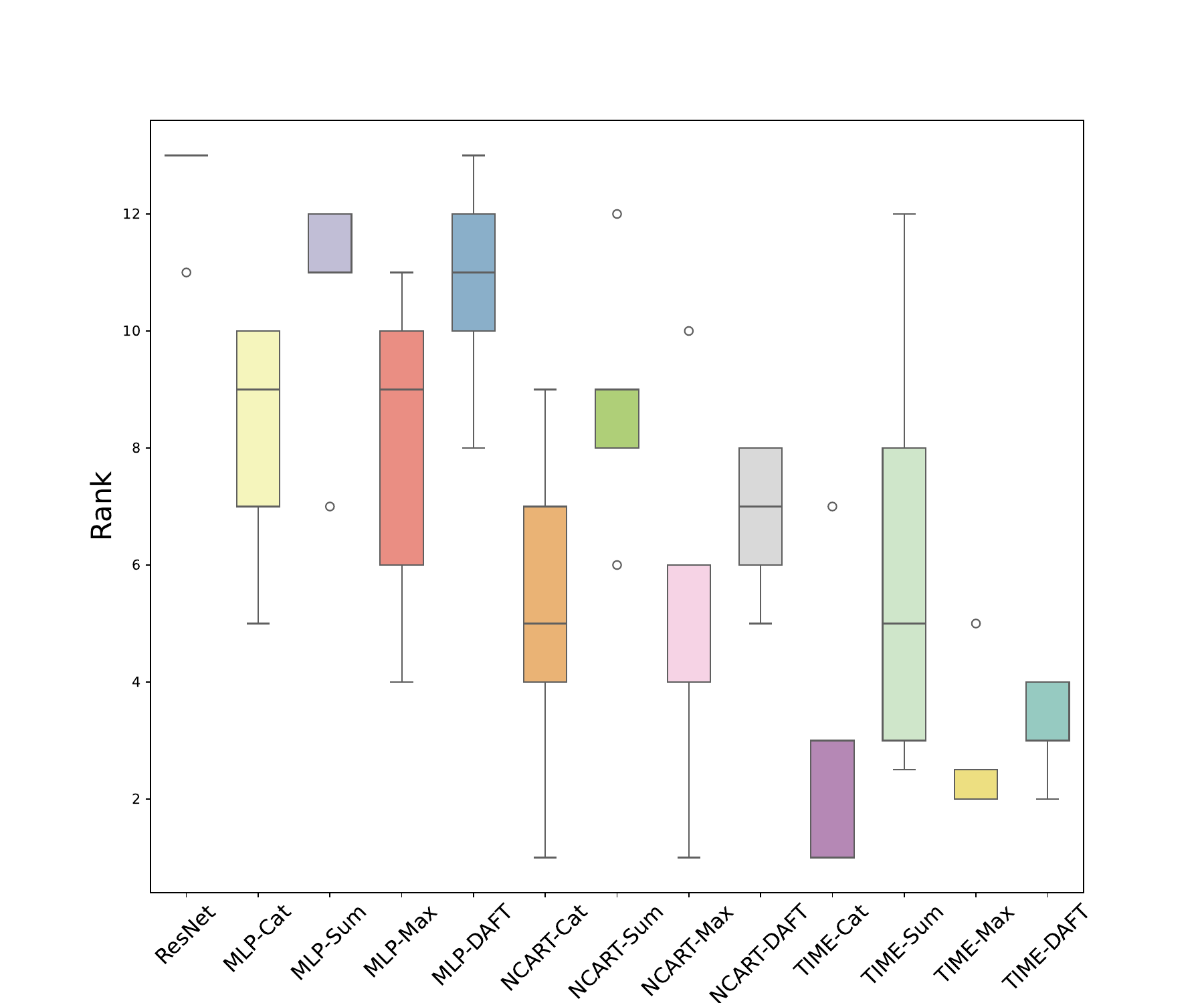}
    \caption{Rankings under the frozen setting on five datasets.}
    \label{f.frozenrank}
\end{figure}

We first impute missing values with the median for the three incomplete medical datasets and then make comparisons on the five complete datasets. Table.~\ref{T.results} presents the performance of various multimodal models across five datasets, evaluating them based on different tabular encoders, fusion strategies, and training settings (frozen vs. fully tuned). Fig.~\ref{f.fullyrank} and \ref{f.frozenrank} show the rankings of different models across the five datasets.

From Table.~\ref{T.results}, multimodal models consistently show significant performance improvements compared to the vision-only ResNet model.
This underscores the benefit of incorporating complementary tabular data into the learning process. Notably, for the natural image dataset Adoption, the frozen ResNet model significantly outperforms its fully tuned counterpart, suggesting that the pretrained vision encoder has already learned highly transferable visual features. A smaller advantage is also observed on the Painting dataset, likely because paintings are often created to depict the real world, allowing pretrained features to generalize reasonably well.
However, on medical image datasets, the frozen ResNet performs worse, likely due to its pretraining on natural image data, which makes it less effective at extracting domain-specific medical features. This highlights the value of using complementary structured data in medical image tasks.

TIME models outperform both NCART and MLP baselines, achieving superior results in terms of both individual best scores and average performance. Notably, TIME leads to significant improvements on all three medical datasets. On the two natural image datasets, where the fully tuned ResNet model fails to surpass the frozen ResNet, integrating tabular information through TIME enables the multimodal model to outperform the frozen baseline. These results highlight the strong generalization ability of TabPFN’s pretrained tabular representation, and demonstrate the effectiveness of leveraging foundation models to enhance multimodal learning across diverse domains.

While TIME models dominate, NCART-based models also show competitive performance on several datasets, outperforming others in certain fusion configurations. For example, NCART-Cat performs well on the Skin Cancer dataset, and NCART-Max is effective on the Painting dataset. However, on the Adoption dataset, the MLP tabular encoder underperforms, with its results remaining close to or below the ResNet model. This suggests that the MLP may struggle to extract meaningful representations from the tabular features of the Adoption dataset, while specialized tabular encoders like NCART—designed to capture tabular feature interactions—can significantly improve performance.

The effectiveness of fusion strategies varies notably across datasets and settings. As shown in Fig.\ref{f.fullyrank} and Fig.\ref{f.frozenrank}, TIME equipped with various fusion strategies consistently achieves stronger results than baseline methods. Although performance slightly declines under the frozen setting, TIME remains competitive, and NCART-based models achieve comparable performance—particularly when combined with the Concatenation and Max fusion strategies.
Overall, Concatenation and DAFT provide robust baselines, performing competitively across datasets. Max fusion demonstrates excellent performance when paired with NCART or TabPFN under the frozen configuration. In contrast, Sum fusion excels when integrated with TabPFN under the fully tuned setting, suggesting that simple additive interactions benefit from the rich, pretrained representations of TabPFN. These results underscore the importance of selecting the most suitable fusion strategy depending on both the dataset and the model, rather than relying on a one-size-fits-all approach.

\subsection{Performance Robustness on Incomplete Data}

\begin{table*}[!ht]
\renewcommand\arraystretch{1.3}
\centering
\caption{Quantitative results (Mean$\pm$std.) of our method on three raw incomplete datasets. The \underline{\textbf{bold}} indicates the best result for each dataset, and the \textit{\textbf{bold}} denotes the best average performance. \textcolor{red}{\ding{88}} indicates full fine-tuning, and \textcolor{blue}{\ding{100}} denotes linear probing (frozen setting). The gray-colored “Mean” line indicates the average scores on incomplete datasets, while the “Mean(imputed)” line represents the average scores on imputed datasets (shown in Table.~\ref{T.results}).}
\label{T.incompleteresults}
\begin{adjustbox}{width=\textwidth}
\begin{tabular}{l|cc|cc|cc}
\toprule[2pt]
\multirow{2}{*}{Models} & \multicolumn{2}{c|}{Breast Cancer (Acc. $\uparrow$)} & \multicolumn{2}{c|}{Covid-19 (Acc. $\uparrow$)} & \multicolumn{2}{c}{Skin Cancer (Acc. $\uparrow$)} \\
  &   \Large\textcolor{red}{\ding{88}}   &   \Large\textcolor{blue}{\ding{100}}   &   \Large\textcolor{red}{\ding{88}}   &   \Large\textcolor{blue}{\ding{100}}  &   \Large\textcolor{red}{\ding{88}}  &   \Large\textcolor{blue}{\ding{100}}   \\

\midrule[1pt]
TIME-Cat  &  72.21\small$\pm$1.52  & \underline{\textbf{73.81\small$\pm$1.20}}    &  \underline{\textbf{92.56\small$\pm$0.89}}      &  89.02\small$\pm$0.77     &  81.74\small$\pm$1.29     &   79.26\small$\pm$0.72       \\

TIME-Sum   &  72.33\small$\pm$1.93   & 72.09\small$\pm$1.80         &  90.12\small$\pm$1.30         &  89.39\small$\pm$1.24       &  80.65\small$\pm$1.14     &   78.65\small$\pm$0.68       \\

TIME-Max  &  \underline{\textbf{73.25\small$\pm$2.66}}   & 72.21\small$\pm$1.30   &  91.83\small$\pm$0.62         &  \underline{\textbf{90.24\small$\pm$0.77}}       &  81.87\small$\pm$1.29         &   79.19\small$\pm$0.56       \\

TIME-DAFT   &  71.96\small$\pm$1.11   & 73.50\small$\pm$0.74       &  91.46\small$\pm$1.93        &  89.88\small$\pm$1.47    &  \underline{\textbf{82.13\small$\pm$0.68}}        &  \underline{\textbf{79.91\small$\pm$1.41}}     \\

\hline
\rowcolor{gray!15}
Mean &  72.44\small$\pm$1.81   & \textit{\textbf{72.90\small$\pm$1.28}}       & \textit{\textbf{91.49\small$\pm$1.18}}    &   \textit{\textbf{89.63\small$\pm$1.11}}   &   \textit{\textbf{81.60\small$\pm$1.03}}    &   \textit{\textbf{79.25\small$\pm$0.84}}   \\

\rowcolor{gray!15}
Mean(imputed)  &  \textit{\textbf{72.70\small$\pm$2.24}}  & 72.70\small$\pm$1.24    &  89.60\small$\pm$1.62        &  87.95\small$\pm$1.01        &  79.40\small$\pm$1.30        &  78.08\small$\pm$0.54      \\

\bottomrule[2pt]
\end{tabular}
\end{adjustbox}
\end{table*}

\begin{figure}[!ht]
    \centering
    \includegraphics[width=\linewidth]{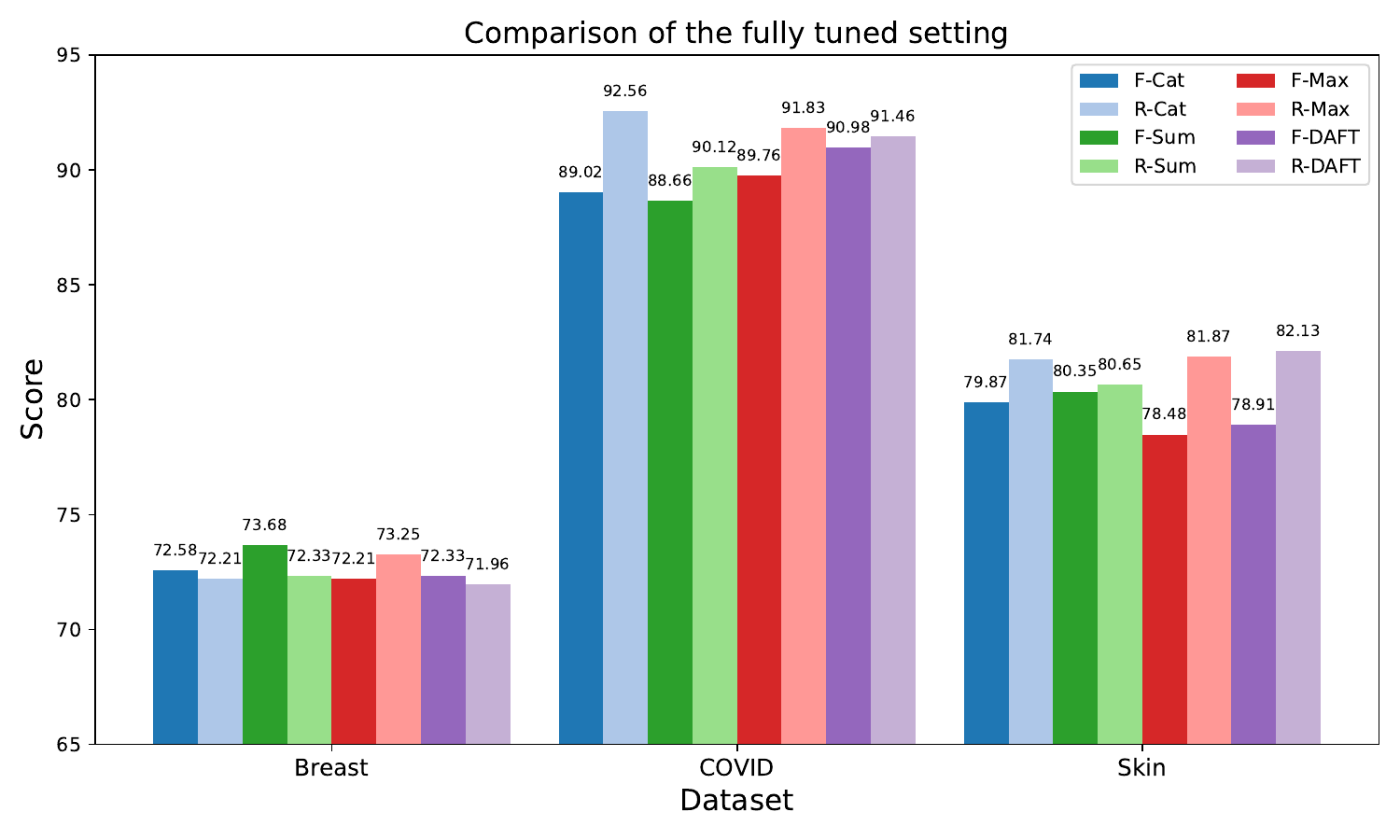}
    \caption{Comparison of the fully tuned setting on three medical datasets. Dark-colored bars (F-) represent results on imputed complete datasets, while light-colored bars (R-) represent results on raw incomplete datasets. Cat, Sum, Max, and DAFT denote the four fusion strategies.}
    \label{f.fully}
\end{figure}

\begin{figure}[!ht]
    \centering
    \includegraphics[width=\linewidth]{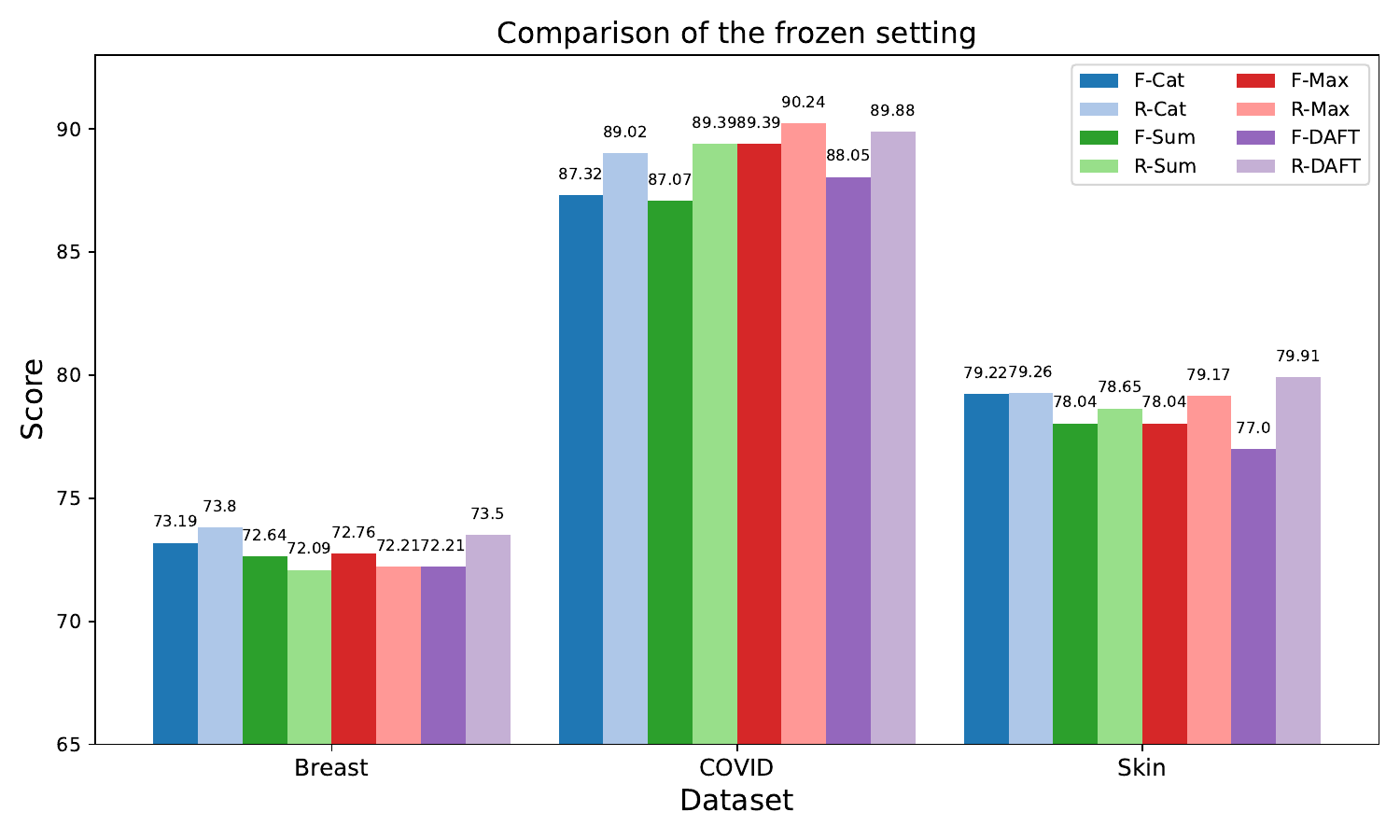}
    \caption{Comparison of the frozen setting on three medical datasets. Dark-colored bars (F-) represent results on imputed complete datasets, while light-colored bars (R-) represent results on raw incomplete datasets. Cat, Sum, Max, and DAFT denote the four fusion strategies.}
    \label{f.frozen}
\end{figure}

We evaluate the robustness of the TIME framework under real-world conditions using three medical datasets that contain missing tabular values. Table.~\ref{T.incompleteresults} presents quantitative results on raw (incomplete) data, while Fig. \ref{f.fully} and \ref{f.frozen} offer a visual comparison between models trained on imputed data and raw incomplete data, under both fully tuned and frozen settings.

Across all datasets and fusion strategies, we observe that models trained directly on raw incomplete data generally achieve higher scores than those trained on median-imputed data. This trend holds in both fully tuned and frozen settings. 
Remarkably, the best mean scores across all tasks are obtained from models trained on raw incomplete datasets.

For instance, F-Cat under the frozen setting achieves the highest accuracy (73.81\%) on Breast Cancer, even outperforming its fully tuned imputed counterpart. On COVID-19, the fully tuned F-Cat model obtains 92.56\%, marking a significant improvement. Although NCART-based models achieve the highest average score (88.29\%) and best individual score (89.76\%) on the imputed COVID-19 data under the frozen setting, the R-Max model (\textit{i.e.}, TIME-Max on the raw data with the frozen setting) surpasses both with a best score of 90.24\% and raises the average to 89.63\%. Similarly, on the Skin Cancer dataset, F-DAFT achieves the best accuracy (82.13\%) under missing data, again outperforming all imputed-data models. 
Referring to the missingness statistics of the COVID-19 dataset in \ref{a.data}, we observe that all samples contain missing values. Notably, across this dataset, all raw-data variants consistently outperform their imputed counterparts by a substantial margin, further underscoring the effectiveness of the TIME framework in handling incomplete tabular data without the need for imputation.

These results highlight an important finding: imputation may introduce noise or distort the underlying data distribution, ultimately degrading model performance. In contrast, TabPFN’s ability to natively handle missing values allows TIME to utilize incomplete data more effectively, without relying on potentially harmful preprocessing.

\subsection{Missingness Sensitivity Analysis}
\begin{figure}[!ht]
    \centering
    \includegraphics[width=\linewidth]{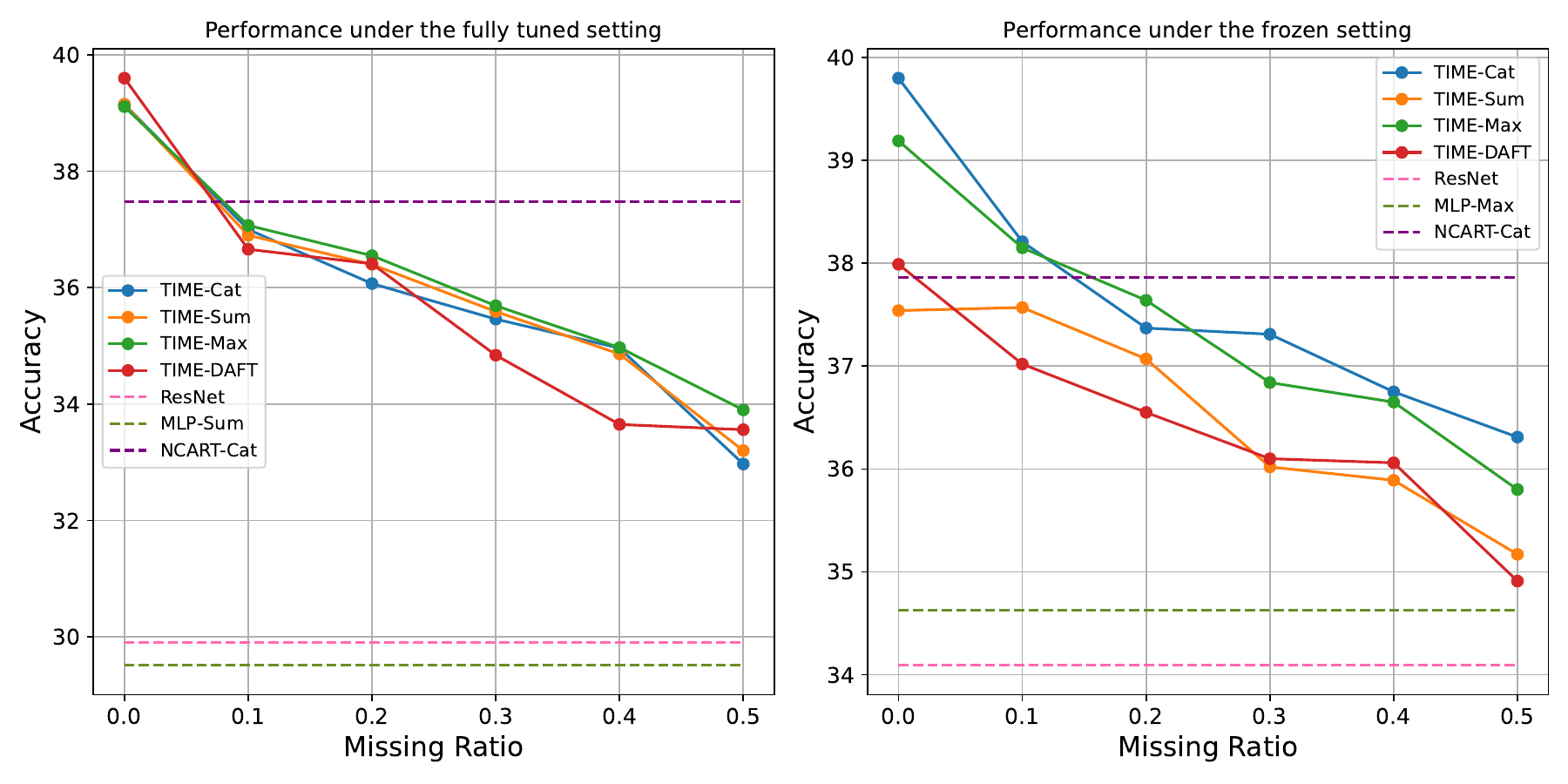}
    \caption{Results of TIME under different masking ratios on the Adoption dataset, evaluated in both fully tuned and frozen settings. The three dotted lines represent the best scores for the three baselines: the pink line corresponds to the ResNet model, the olive line to the best MLP-based model, and the purple line to the best NCART-based model.}
    \label{f.sensitivity}
\end{figure}

To evaluate the robustness of our method under varying levels of missing values, we conduct a missingness sensitivity analysis on the adoption dataset. We randomly mask values in the tabular modality at five levels: 10\%, 20\%, 30\%, 40\%,  and 50\%. Since other baseline methods cannot operate directly on incomplete data, we focus exclusively on our proposed TIME framework and report classification accuracy at each missingness level across different fusion strategies under both frozen and fully tuned settings.

The results, summarized in Fig.~\ref{f.sensitivity}, demonstrate that the performance of TIME remains relatively stable as the proportion of missing data increases. Under the fully tuned setting, we observe a more noticeable performance drop compared to the frozen setting. This is expected, as the fully tuned model relies more heavily on the complete tabular input to enhance prediction. When this information becomes incomplete, the model’s effectiveness is partially compromised. Nonetheless, TIME still outperforms the best results achieved by MLP-based (MLP-Cat) and ResNet models, even with significant missingness.

In contrast, under the frozen setting, the model shows reduced sensitivity to missing tabular data. Since the image encoder already provides strong representations, the model depends less on detailed tabular patterns, resulting in a more modest performance decline. Nevertheless, the frozen version of TIME still outperforms both the MLP-based (MLP-Max) and ResNet baselines. Although the NCART-based model performs well on imputed datasets, it lacks native support for missing values and therefore cannot be applied in this setting.

While accuracy gradually decreases with higher levels of missingness, the degradation is not severe—indicating that TIME is robust to missing values. Notably, fusion strategies such as Concatenation and Element-wise Max show greater resilience across datasets. This suggests that these fusion mechanisms are better able to preserve and integrate the partial information available from both modalities under varying degrees of incompleteness.

\section{Conclusions}

In this paper, we addressed two fundamental challenges in tabular-image multimodal learning: the absence of a standardized pretrained representation for tabular data and the lack of robustness to missing values—both of which are especially critical in applications. To overcome these issues, we proposed TIME, a simple yet effective multimodal learning framework that integrates the pretrained TabPFN model as a frozen tabular encoder and combines its output with image features from a vision backbone.

TIME brings the strengths of foundation models—robustness, transferability, and zero-shot generalization—into the tabular modality. Crucially, it natively accommodates incomplete data without requiring imputation. Through comprehensive experiments on both natural and clinical datasets, we demonstrated that TIME consistently outperforms existing baselines across multiple fusion strategies and under both complete and missing-value settings. These findings highlight the potential of pretrained tabular encoders to substantially enhance multimodal learning in terms of performance, generalization, and robustness.

Despite these promising results, several limitations remain. First, the applicability of TIME is constrained by the current limitations of TabPFN, which supports only datasets with fewer than 10000 samples, 500 features, and 10 classes \cite{hollmann2025accurate}, making it suitable primarily for small- to medium-sized datasets. Second, the fusion strategies used in this work are relatively simple and may be further improved. In future work, we will focus on scaling the approach to larger datasets, exploring more advanced and adaptive fusion techniques and enhancing the interpretability of multimodal predictions, especially in high-stakes domains like healthcare, where trust and transparency are essential.

\section*{Acknowledgment}
This work was partially supported by the National Natural Science Foundation of China no. 12271492.

\appendix
\section{Datasets Description}
\label{a.data}

Five datasets are used in the experiments:
\begin{itemize}
    \item \textbf{Adoption} is a classification task that predicts the speed at which a pet is adopted. The image modality consists of pet photos, while the tabular data includes attributes such as pet type and health status.
    \item \textbf{Breast Cancer} is a classification task that predicts the cancer stage. The images are mammograms, and the tabular features include variables like breast density and pathology.
    \item \textbf{Covid-19} is a classification task to determine whether a patient has COVID-19 or another respiratory disease. The image data consists of chest X-rays, while the tabular features include temperature, age, and other clinical indicators.
    \item \textbf{Skin Cancer} is a classification task that identifies the category of skin lesions. The image modality comprises dermoscopic images, and the tabular data contains relevant medical history, such as smoking and cancer history.
    \item \textbf{Painting} is a regression task that predicts the price of a painting. The image input is the painting itself, and the tabular data includes metadata such as artistic style (\textit{e.g.}, realist, expressionist) and other relevant attributes.
\end{itemize}

\begin{table}[!ht]
\renewcommand\arraystretch{1.3}
\centering
\caption{Dataset details. \#Samples indicates the total number of samples in the dataset; \#Features refers to the number of tabular features; \#Classes denotes the number of target classes; \#Missingness represents the number of samples containing missing values; \#Train/Val/Test shows the number of samples in the training, validation, and test sets, respectively.}
\label{T.data}
\begin{adjustbox}{width=\textwidth}
\begin{tabular}{c|c|c|c|c|c|c}
\toprule[2pt]
Dataset & Image type  & \#Sample  & \#Feature & \#Class & \#Missingness & \#Train/Val/Test \\
\midrule[1pt]
Adoption & natural & 14652 &  19 & 5 & 0 & 9376/2345/2931 \\
 \hline
Breast Cancer & medical & 1870 & 8 & 3 & 462 & 1235/309/326 \\
 \hline
Covid-19 & medical & 820 & 17 & 3 & 820 & 524/132/164 \\
 \hline
Skin Cancer & medical & 2298 & 22 & 6 & 824 & 1470/368/460 \\
 \hline
Painting & painting & 12369 & 245 & 1 & 0 & 7916/1979/2474 \\
 \bottomrule[2pt]
\end{tabular}
\end{adjustbox}
\end{table}

\bibliographystyle{elsarticle-num} 
\bibliography{references}

\end{document}